\documentclass[lettersize,journal]{IEEEtran}
\usepackage{amsmath,amsfonts}
\usepackage{algorithmic}
\usepackage{algorithm}
\usepackage{array}
\usepackage[caption=false,font=normalsize,labelfont=sf,textfont=sf]{subfig}
\usepackage{textcomp}
\usepackage{stfloats}
\usepackage{url}
\usepackage{verbatim}
\usepackage{booktabs}
\usepackage{multirow}
\usepackage{cite}
\usepackage{makecell}
\usepackage{graphicx}
\begin{document}

\title{SED-FOD: Scattering-Aware Expert Decomposition for Few-Shot Cross-Sensor SAR Object Detection}

\author{
Shu Yang, Zhen Chen, Zhiyu Jiang, Yanlei Li, and Xingdong Liang
\thanks{
The authors are with the National Key Laboratory of Microwave Imaging
Technology, Aerospace Information Research Institute,
Chinese Academy of Sciences, Beijing 100190, China, and also with the
School of Electronic, Electrical and Communication Engineering,
University of Chinese Academy of Sciences, Beijing 100049, China.
}%
\thanks{
Corresponding author: Xingdong Liang
(e-mail: liangxd@aircas.ac.cn).
}%
\thanks{This work has been submitted to IEEE for possible publication.
Copyright may be transferred without notice, after which this version
may no longer be accessible.}%
}

\maketitle

\begin{abstract}
Synthetic aperture radar (SAR) object detection is an important part of remote sensing interpretation. However, because of variations in frequency band, resolution, background clutter, and target scattering responses, the performance of existing detectors often degrades when training and testing data are acquired from different SAR domains. Although domain adaptation methods offer a promising paradigm for solving this problem, most of them mainly pursue domain-invariant feature alignment and suppress sensor-dependent scattering characteristics that are useful for object detection. This problem becomes more challenging in few-shot scenarios, where only a few fully annotated target-domain SAR images are available. To address this issue, we propose a scattering-aware shared-specific feature decomposition framework for few-shot SAR domain adaptation object detection. We decompose detection features into a shared path and several soft-gated scattering-specific expert paths. The shared path learns transferable object structural information and is used for asymmetric domain alignment, while the scattering-specific experts adaptively compensate heterogeneous SAR responses. In addition, routing-domain auxiliary loss is introduced to encourage specific experts to capture sensor-dependent routing preferences, and an expert balancing loss is used to prevent routing collapse. Extensive experiments on four bidirectional heterogeneous SAR detection tasks between FARAD-X/FARAD-Ka and MiniSAR under different few-shot settings have been conducted and experimental results demonstrate that the proposed method achieves superior performance in both forward and
reverse adaptation directions.
\end{abstract}

\begin{IEEEkeywords}
Synthetic aperture radar (SAR) object detection, few-shot domain adaptation, cross-sensor, mixture of experts.
\end{IEEEkeywords}

\section{Introduction}

Synthetic Aperture Radar (SAR) is an active microwave imaging sensor that utilizes pulse compression and synthetic aperture techniques to provide two-dimensional high-resolution imagery. Owing to its all-day and all-weather capabilities, SAR is widely used in remote-sensing interpretation tasks~\cite{shi2021unsupervised,Kong2025class,Li2017ship}.
SAR object detection is an essential task in SAR image interpretation for locating targets of interest, such as vehicles, ships, and aircraft, from complex backgrounds~\cite{SARATR-X,HRSID,zou2022vehicle}. 

In recent years, deep-learning-based detectors have significantly improved SAR object detection performance by learning discriminative representations from large amounts of annotated data with similar statistical characteristics~\cite{chen2020learning, zhang2021dense,huang2023evidential}. 
However, most existing SAR image object detection studies train and evaluate their model under homogeneous imaging conditions, where training and testing images are acquired from the same or similar sensors. 
When evaluated on images collected under different SAR imaging conditions,
their performance may degrade significantly because of cross-sensor
distribution discrepancies.~\cite{Xu2022Eyes}. 
In practical applications, SAR images are often collected by different sensors with diverse frequency bands, resolutions, imaging geometries, polarization modes, and background environments. These factors lead to significant differences in target characteristics and feature distributions~\cite{zhang2025cross}. 
Consequently, the same vehicle target may present obviously different signatures across sensors, while background structures may also produce confusing scattering patterns leading to degraded detection performance and reduced generalization ability. 

Unlike conventional transfer learning methods, which usually require sufficient labeled target-domain data from target domain to finetune the model, heterogeneous SAR object detection often suffers from very limited labeled target-domain data. As a result, the distribution discrepancy between training and testing data cannot be effectively overcome~\cite{zhao2022transferable}.
To solve this problem, domain adaptation methods have been proposed. These methods usually reduce the distribution discrepancy between the source and target domains by mapping the data from the source domain and the target domain into the same feature space and aligning domain-invariant representations~\cite{wang2018deep,wilson2020survey,singhal2023domain}. Domain adaptation has been proven to be an effective approach to object detection tasks and is currently widely applied in the fields of computer vision and optical remote sensing.
\begin{figure}[t]
    \centering
    \includegraphics[width=1\linewidth]{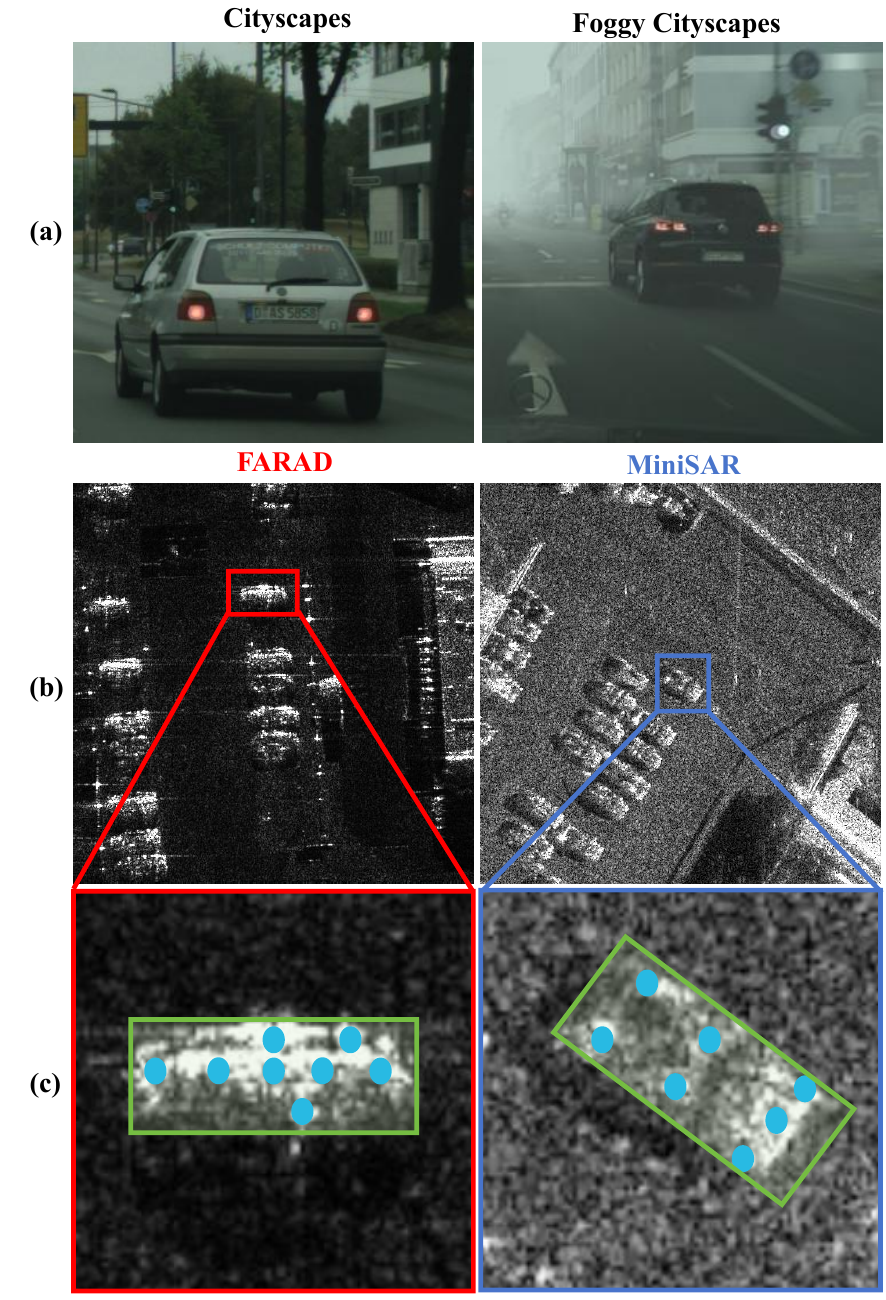}
    \caption{
    Comparison between optical and cross-sensor SAR domain shifts.
    (a) Cityscapes and Foggy Cityscapes illustrate low-level visual variations caused by fog, which provide limited target-discriminative cues.
    (b) FARAD and MiniSAR show cross-sensor SAR discrepancies in clutter and target scattering responses.
    (c) Enlarged vehicle patches highlight shared target structures with green boxes and domain-specific scattering cues with blue dots.
    }
    \label{fig:motivation}
\end{figure}
In particular, most previous studies focus on unsupervised domain adaptation, which requires abundant unlabeled data from the target domain. Although this can reduce the high cost caused by the fact that SAR image interpretation usually requires expert knowledge such a requirement is impractical when target data are inherently scarce. This scenario often occurs when a new SAR platform captures a new batch of data and requires rapid preliminary interpretation. Therefore, developing a few-shot domain adaptation strategy is crucial for practical cross-sensor SAR object detection~\cite{ZHAO2025100202}.

However, few-shot domain adaptation for SAR object detection still faces certain challenges. Most existing studies mainly pursue domain-invariant representation learning by aligning the source and target feature distributions~\cite{gao2023asyfod,kumar2018co,zhang2024few}. This alignment strategy has been widely adopted in optical domain adaptation, where domain shifts are often dominated by low-level appearance variations such as illumination, weather and styles. As shown in Fig.~\ref{fig:motivation}(a), the clear-to-foggy variation in optical images (Foggy Cityscapes~\cite{sakaridis2018semantic}) mainly changes low-level appearance, while the vehicle structure and scene layout remain largely consistent. Such domain-specific visual variations usually provide limited direct cues for target discrimination and are often treated as nuisance factors to be aligned or suppressed. In contrast, cross-sensor SAR vehicle detection presents a more challenging scenario. Fig.~\ref{fig:motivation}(b) shows that cross-sensor SAR images exhibit significant discrepancies in background clutter and target scattering responses. The enlarged vehicle patches in Fig.~\ref{fig:motivation}(c) further indicate that SAR targets across domains still share certain structural cues, while their local scattering responses vary noticeably. Therefore, the domain discrepancy in SAR imagery should not be regarded as a simple nuisance appearance shift, and directly enforcing the entire feature representation to be domain-invariant may be suboptimal.~\cite{shi2024unsupervised,decom2022zhao}. 

This makes existing feature-level alignment methods potentially problematic from two aspects.
On the one hand, domain-specific features like scattering centers, shadow responses, and resolution-dependent target structures may be weakened if incorrectly treated as noise, which reduces target objectness and leads to missed detections. 
On the other hand, background structures such as building edges, parking lots, sidelobe-like bright points, and clutter may exhibit vehicle-like local scattering responses. Forcing source and target features to be aligned through directly minimizing domain discrepancy with only a few target samples may pull such confusing background responses toward the target feature space, resulting in false alarms. 
Therefore, the key challenge of few-shot SAR adaptation is not simply to remove domain-specific information, but to distinguish transferable target structures from specific discriminative information that is still useful for detection.

This observation motivates us to rethink the role of domain-specific discriminative information in few-shot SAR adaptation. 
Instead of simply  minimizing domain discrepancy, we argue that the detection representation should be decomposed into transferable shared features and sensor-dependent scattering-specific characteristics. 
The domain invariant representation is expected to capture domain-transferable target structures while the scattering-specific representation should be preserved and adaptively exploited to compensate for heterogeneous SAR responses.

To overcome this issue, we propose a scattering-aware shared-specific feature decomposition framework for few-shot SAR domain adaptive object detection. The proposed method introduces a shared path and multiple soft-gated specific expert paths. The shared path is used to learn transferable target representations to guide subsequent source data selection and domain alignment while the scattering-specific representations learned by the experts will be adaptively weighted to model specific characteristics and provide compensation for object prediction. 
With the help of both paths, the proposed method aligns transferable structures and retains important scattering-specific information to reduce the risk of over-alignment and over-fitting in few-shot heterogeneous SAR adaptation.

The main contributions are as follows:
\begin{itemize}
    \item We propose a scattering-aware shared-specific feature decomposition framework for few-shot SAR domain adaptive object detection, which separates transferable domain invariant representation from sensor-dependent scattering characteristics.
    \item We design a soft gated scattering-specific expert compensation module, where multiple experts adaptively model heterogeneous SAR scattering variations.
    \item We perform source data selection and domain alignment on the shared representation, reducing the risk of over-aligning and over-fitting. Extensive experiments demonstrate that the proposed method significantly outperforms state-of-the-art methods.
\end{itemize}

\section{Related Work}
\label{sec:related_work}

\subsection{SAR Object Detection and Domain Adaptation}

SAR object detection has received increasing attention because SAR sensors provide reliable imaging under diverse weather and illumination conditions~\cite{li2024sardet}. Early methods mainly relied on statistical modeling and handcrafted features, among which constant false alarm rate methods and their variants were extensively studied~\cite{weiss1982analysis,frery1997model,tison2004new}. With the development of deep learning, CNN based detectors have substantially improved detection performance in complex SAR scenes~\cite{wang2018sar,krizhevsky2012imagenet,kechagias2021automatic}. However, SAR imagery differs fundamentally from optical imagery and is characterized by sparse scattering centers, target shadows, speckle noise, and strong background clutter~\cite{oliver2004understanding,jakowatz2012spotlight,HUANG20091533}. Target responses also vary with frequency band, spatial resolution, imaging geometry, polarization, and background conditions~\cite{ZHAO2025100202}. Consequently, detectors trained under homogeneous imaging conditions often generalize poorly to heterogeneous SAR data, particularly when only limited target domain annotations are available~\cite{zhou2024conditional}.

Domain adaptive object detection addresses this problem by reducing the distribution discrepancy between source and target domains~\cite{li2024comprehensive}. Feature alignment methods align image level or instance level representations through adversarial learning or progressive adaptation~\cite{chen2018domain,guo2021sar,zou2023cross}. Data enhancement methods instead generate target like samples or combine source and target content, as exemplified by transition image generation, Cross Domain CutMix, and Fourier domain adaptation~\cite{shi2021unsupervised,Nakamura2022FewshotAO,Yang2020FDAFD}. Another line of work adopts pseudo labeling and teacher student learning to progressively refine target domain supervision~\cite{Cai2019ExploringOR,ramamonjison2021simrod,Zhou2022SSDAYOLOSD,glenn_jocher_2022_734952}. Although effective, these methods typically require sufficient target samples to estimate the target distribution or construct reliable pseudo labels. Their applicability is therefore limited when a newly deployed SAR platform provides only a small number of annotated images.

\subsection{Few Shot Cross Sensor SAR Object Detection}

Few shot object detection adapts a detector to novel scenarios using only a small set of annotated samples~\cite{Song2022ACS}. A common strategy first trains the detector on source data and then fine tunes selected components using limited target samples. FSDet demonstrates that simple fine tuning provides a competitive baseline~\cite{Wang2020FrustratinglySF}, while AcroFOD improves adaptation through domain aware data enhancement and candidate filtering~\cite{Gao2022AcroFODAA}. AsyFOD further introduces target guided source selection and asymmetric alignment to address the imbalance between abundant source samples and scarce target samples~\cite{gao2023asyfod}. For SAR imagery, FsDAOD employs mutual information regularization to preserve domain related knowledge~\cite{ZHAO2025100202}, while CS FSDet introduces distribution alignment and domain interaction modules for cross sensor adaptation~\cite{liu2025cs}. These methods demonstrate the value of few shot adaptation when both domain discrepancy and target data scarcity are present.

Most existing approaches nevertheless emphasize domain invariant representation learning through feature alignment or enhancement. This strategy may be suboptimal for heterogeneous SAR imagery because sensor dependent variations contain not only nuisance factors but also discriminative scattering cues. Enforcing complete invariance may suppress useful scattering centers, shadow structures, and resolution dependent responses, thereby causing negative transfer. More broadly, conditional representation learning has shown that explicitly preserving task relevant conditions can improve controllability and consistency in pose guided generation, virtual dressing, and story visualization~\cite{shen2024imagpose,shen2024advancing,shen2025imagdressing,shen2025boosting}. Motivated by this principle, our method decomposes the detection representation into transferable shared features and sensor dependent scattering features. Domain adaptation is performed only on the shared representation, while scattering specific experts preserve and adaptively compensate for heterogeneous SAR responses.
 
\section{Method}
\begin{figure*}[t]
    \centering
    \includegraphics[width=1\textwidth]{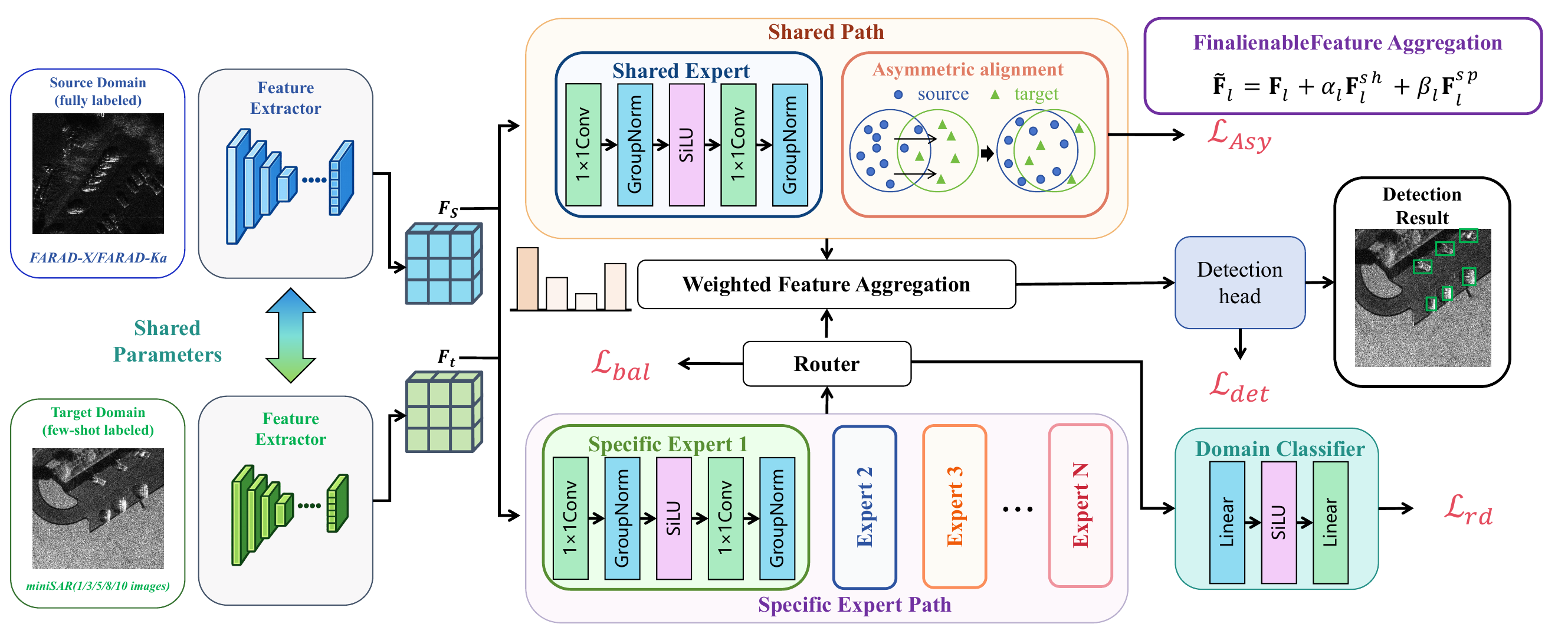}
    \caption{
    Overall framework of the proposed method. 
    Source and few-shot target SAR images are fed into a shared feature extractor, followed by an MoE feature learning module composed of an always-on shared path and router-weighted scattering-specific experts. 
    The shared representation is used for asymmetric domain alignment, while the aggregated expert features provide domain-sensitive compensation for object detection.
    }
    \label{fig:pipeline}
\end{figure*}
The overall framework of the proposed method is illustrated as Fig.~\ref{fig:pipeline}. 
Following the asymmetric Adaptation Paradigm~\cite{gao2023asyfod}, source and few-shot target SAR images are first fed into the backbone to extract hierarchical features, which are then used for source selection, domain alignment, and object detection.
However, directly applying selection and alignment to the whole detection feature may over-align sensor-dependent scattering responses in cross-sensor SAR imagery.
To address this issue, we decompose the detection feature into an always-on shared path and multiple soft-gated scattering-specific expert paths.
The shared representation is used for source selection and asymmetric alignment, while the scattering-specific experts provide adaptive compensation for object detection.

The remainder of this section is organized as follows. Section~\ref{subsec:problem_formulation} formulates the few-shot cross-sensor SAR object detection problem. Section~\ref{subsec:feature_decomposition} introduces the proposed shared-specific feature decomposition. Section~\ref{subsec:scattering_experts} presents the soft-gated scattering-specific expert compensation module. Section~\ref{subsec:shared_adaptation} describes shared-representation-guided source selection and domain alignment. Finally, Section~\ref{subsec:objective} gives the overall optimization objective.

\subsection{Problem Formulation}
\label{subsec:problem_formulation}

Let $\mathcal{D}_{s}=\{(x_i^s,y_i^s)\}_{i=1}^{N_s}$ be the labeled source-domain SAR dataset, where $x_i^s$ denotes a source SAR image and $y_i^s=\{(b_{ij}^s,c_{ij}^s)\}_{j=1}^{M_i^s}$ denotes the corresponding object annotations, including bounding boxes $b_{ij}^s$ and class labels $c_{ij}^s$. The target-domain few-shot training set is denoted as $\mathcal{D}_{t}^{K}=\{{(x_i^t,y_i^t)}\}_{i=1}^{K},$ where only $K$ fully annotated target-domain images are available for adaptation. 
In this work, $K$ denotes the number of fully annotated target-domain images rather than the number of object instances. The source and target domains share the same object category but are collected by different scenes, leading to a significant distribution discrepancy, i.e. $P_s(x,y) \neq P_t(x,y)$.

\subsection{Scattering-Aware Shared-Specific Feature Decomposition}
\label{subsec:feature_decomposition}

The above analysis indicates that few-shot cross-sensor SAR adaptation should not only reduce the source--target distribution discrepancy, but also handle sensor-dependent scattering variations. Therefore, the feature representation is expected to satisfy two coupled objectives. First, the detector should align transferable object structures so that vehicle-related representations can be shared across sensors, which helps reduce missed detections caused by severe domain shift. Second, the detector should preserve and adaptively exploit sensor-dependent scattering responses, so that useful target scattering cues are not suppressed and confusing background scatterers can be better distinguished from real vehicles. To this end, we decompose the detection feature into a shared component for domain adaptation and scattering-specific expert components for residual compensation.

Given an intermediate detection feature map $\mathbf{F}_{l}\in\mathbb{R}^{C_l\times H_l\times W_l}$ at the $l$-th feature level, we decompose it into a domain-shared representation and multiple scattering-specific representations. The domain-shared path is
designed to extract transferable responses from the contextual features encoded by the detector, including sensor-stable cues related to target structure, spatial configuration, and object-background relations. It is formulated as
\begin{equation}
\mathbf{F}_{l}^{\mathrm{sh}}=G_{l}^{\mathrm{sh}}(\mathbf{F}_{l}),
\end{equation}
where $G_{l}^{\mathrm{sh}}(\cdot)$ denotes the shared feature transformation.

In parallel, multiple scattering-specific experts are introduced to capture complementary residual responses. The output of the $e$-th expert is given by
\begin{equation}
\mathbf{F}_{l,e}^{\mathrm{sp}}=G_{l,e}^{\mathrm{sp}}(\mathbf{F}_{l}),\qquad e=1,\ldots,E,
\end{equation}
where $E$ is the number of specific experts and $G_{l,e}^{\mathrm{sp}}(\cdot)$ denotes the transformation of the $e$-th expert. Unlike the shared path, the specific expert paths are not explicitly constrained to produce domain-invariant representations. With independent parameters, they can learn complementary channel transformations for scattering-related residual responses associated with variations in frequency band, resolution, clutter, shadow, and scattering-center distribution.

The shared transformation and all specific expert transformations adopt the same lightweight architecture while maintaining independent parameters. Specifically, each transformation consists of two pointwise convolutions with channel expansion and projection:
\begin{equation}
G(\mathbf{F}_{l})=\operatorname{GN}_{2}\left(\mathbf{W}_{2}^{1\times1}*\phi\left(
\operatorname{GN}_{1}\left(\mathbf{W}_{1}^{1\times1}*\mathbf{F}_{l}\right)\right)\right),
\label{eq:expert_transform}
\end{equation}
where $\mathbf{W}_{1}^{1\times1}$ expands the channel dimension from $C_l$ to $rC_l$, $\mathbf{W}_{2}^{1\times1}$ projects it back to $C_l$, $\operatorname{GN}(\cdot)$ denotes group normalization, and $\phi(\cdot)$ is the SiLU activation function. We set the expansion ratio to $r=2$. Using a homogeneous architecture ensures that the complementary behaviors of different experts arise from their independently learned parameters and adaptive routing weights, rather than manually assigned expert structures.

\subsection{Soft-Gated Scattering-Specific Expert Compensation}
\label{subsec:scattering_experts}

Although the shared path provides transferable object structures, a single domain-specific branch is insufficient to represent the heterogeneous scattering variations in cross-sensor SAR images. The sensor-dependent discrepancy may be caused by different factors, such as scattering-center distribution, shadow response, background clutter, resolution degradation, and sidelobe-like interference. These factors vary across images and local scenes, making the domain-specific residual a mixture of multiple scattering patterns rather than a single deterministic transformation.

To model such heterogeneous SAR responses, inspired by the Mixture of Experts model~\cite{jacobs1991adaptive}, we introduce a soft-gated expert compensation mechanism. For the $l$-th feature level, a routing function predicts the importance of each scattering-specific expert according to the input feature:
\begin{equation}
\mathbf{r}_{l}=\mathrm{softmax}\left(\frac{R_l(\mathrm{GAP}(\mathbf{F}_{l}))}{\tau}\right),
\end{equation}
where $R_l(\cdot)$ denotes the routing function, $\mathrm{GAP}(\cdot)$ is global average pooling, $\tau$ is a temperature parameter, and $\mathbf{r}_{l}=[r_{l,1},\ldots,r_{l,E}]$ represents the soft expert weights. The scattering-specific compensation feature is then obtained by
\begin{equation}
\mathbf{F}_{l}^{sp}=\sum_{e=1}^{E} r_{l,e}\mathbf{F}_{l,e}^{sp}.
\end{equation}

To encourage the router to capture domain-related scattering preferences, we further introduce a routing-domain classifier. Given the domain label $d_i\in\{0,1\}$, where $0$ and $1$ denote the source and target domains, respectively, the routing weights from selected feature levels are concatenated as
\begin{equation}
\mathbf{q}_{i}=\mathrm{Concat}\left(\mathbf{r}_{l_1}(x_i),\mathbf{r}_{l_2}(x_i),\ldots,\mathbf{r}_{l_m}(x_i)\right),
\end{equation}
where $\{l_1,\ldots,l_m\}$ denotes the set of feature levels equipped with scattering-specific experts. A domain classifier $C_r(\cdot)$ predicts the domain probability from the routing vector:
\begin{equation}
\hat{\mathbf{d}}_{i}=C_r(\mathbf{q}_{i}).
\end{equation}
The routing-domain supervision is defined as
\begin{equation}
\mathcal{L}_{rd}=\frac{1}{|\mathcal{B}|}\sum_{i\in\mathcal{B}}\mathrm{CE}\left(\hat{\mathbf{d}}_{i}, d_i\right),
\label{eq:routing_domain_loss}
\end{equation}
where $\mathrm{CE}(\cdot)$ denotes the cross-entropy loss and $\mathcal{B}$ is the training mini-batch. This auxiliary loss is not used to make the whole representation domain-invariant. Instead, it encourages the router to softly adjust the weights of scattering-specific experts according to sensor-dependent response patterns. The classifier is used only during training and is removed during inference.

The final output feature is formulated as
\begin{equation}
\widetilde{\mathbf{F}}_{l}=\mathbf{F}_{l}+\alpha_l \mathbf{F}_{l}^{sh}+\beta_l\mathbf{F}_{l}^{sp},
\label{eq:ssmoe_output}
\end{equation}
where $\alpha_l$ and $\beta_l$ control the contributions of the shared representation and the scattering-specific compensation, respectively. The original feature $\mathbf{F}_{l}$ is preserved through a residual connection to maintain the basic detection representation. The shared path is mandatory for all samples and provides a stable cross-sensor feature basis, while the soft-gated scattering-specific experts adaptively compensate for sensor-dependent SAR responses.

It is worth noting that the shared path does not participate in the expert competition. This design is particularly important in the few-shot target setting, where completely relying on routed experts may overfit the limited target-domain samples. By enforcing all samples to pass through the shared path, the detector maintains transferable object representations, while the expert paths only provide adaptive residual compensation for scattering-specific variations.

\subsection{Shared-Representation-Guided Domain Adaptation}
\label{subsec:shared_adaptation}

\begin{figure}[t]
    \centering
    \includegraphics[width=1\linewidth]{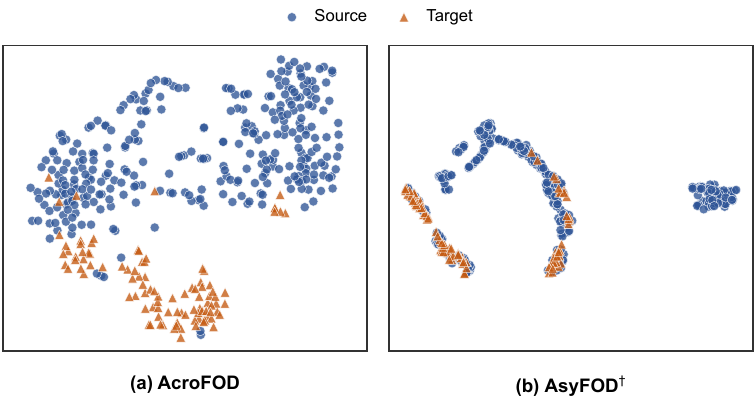}
    \caption{
    Motivating t-SNE visualization of source and target feature distributions under the 1-shot FARAD-Ka $\rightarrow$ MiniSAR setting.
    AcroFOD shows limited source--target overlap, while AsyFOD produces a more concentrated target cluster but still leaves many source samples poorly matched.
    }

    \label{fig:motivation_tsne}
\end{figure}

After obtaining the shared and scattering-specific representations, domain adaptation is performed only on the shared representation. This design is motivated by two observations in few-shot cross-sensor SAR adaptation in Fig.~\ref{fig:motivation_tsne}. First, insufficient feature alignment may leave a large discrepancy between the source and target domains. In this case, the source and target features are scattered in the representation space with only limited overlap, making it difficult for the detector trained on source samples to generalize to the target sensor. Second, directly enforcing strong alignment on the entire detection feature is also risky under the few-shot target setting. Since the target distribution is estimated from only a few annotated target images, the alignment anchor can be severely biased toward limited target scattering patterns. As a result, the target features may become over-concentrated around a small number of target samples, while many source samples still remain poorly matched to the target domain. Conceptually, both cases lead to a low overlap between the source and target distributions compared with their overall feature spread.

Therefore, the key is not to avoid alignment, but to constrain alignment to the transferable part of the representation. The shared path is expected to encode object structures that are relatively stable across sensors, such as target shape, spatial layout, and object-background configuration. In contrast, the scattering-specific expert paths contain sensor-dependent responses related to scattering centers, shadow, clutter, and resolution-induced structures. If the entire fused feature is directly aligned, useful scattering-specific cues may be incorrectly suppressed. Therefore, both target-guided source selection and asymmetric feature alignment are conducted in the shared feature space, while the scattering-specific expert paths are preserved for adaptive compensation.

Specifically, for each target image $x_i^t$, its shared feature at the selected high-level feature layer is extracted and globally pooled as
\begin{equation}
\mathbf{z}_i^t=\mathrm{GAP}\left(\mathbf{F}^{sh}(x_i^t)\right).
\end{equation}
The few-shot target samples are then used to estimate the target-domain shared representation:
\begin{equation}
\bar{\mathbf{z}}_t=\frac{1}{K}\sum_{i=1}^{K}\mathbf{z}_i^t,
\label{eq:target_prototype}
\end{equation}
where $K$ denotes the number of labeled target-domain images available for adaptation.

Similarly, for each source image $x_i^s$, we obtain its shared representation:
\begin{equation}
\mathbf{z}_i^s=\mathrm{GAP}\left(\mathbf{F}^{sh}(x_i^s)\right).
\end{equation}
Following the target-guided source selection strategy, the relevance between each source image and the few-shot target domain is measured in the shared feature space:
\begin{equation}
d_i=\mathcal{D}\left(\mathbf{z}_i^s,\bar{\mathbf{z}}_t\right),
\label{eq:source_distance}
\end{equation}
where $\mathcal{D}(\cdot)$ denotes the feature discrepancy measure. In our implementation, $\mathcal{D}(\cdot)$ is instantiated by the MMD-based discrepancy used in the target-guided source selection procedure. Source samples with smaller discrepancies are considered more relevant to the target domain and are preferentially selected for training~\cite{Gao2022AcroFODAA}. Vitally, source selection is performed on the shared representation rather than the original detector feature. Therefore, the selected source samples are expected to share transferable object structures with the target domain instead of being selected according to sensor-specific scattering responses.

After obtaining the target-similar source samples, asymmetric feature alignment is imposed on the shared representation~\cite{gao2023asyfod}. 
Given a mini-batch containing selected source samples and few-shot target samples, the shared alignment loss is formulated as
\begin{equation}
\mathcal{L}_{align}^{sh}=\mathcal{A}_{asy}\left(\{\mathbf{z}_i^s\}_{i=1}^{n_s},\{\mathbf{z}_j^t\}_{j=1}^{n_t}\right),
\label{eq:shared_alignment}
\end{equation}
where $\mathcal{A}_{asy}(\cdot)$ denotes the adopted asymmetric distribution alignment criterion, and $n_s$ and $n_t$ are the numbers of selected source samples and target samples in the mini-batch, respectively. Different from symmetric alignment that treats the two domains equally, the asymmetric alignment strategy uses the limited target-domain samples as adaptation anchors and encourages the selected source shared features to approach the target-domain distribution. In our implementation, the original asymmetric alignment criterion is applied to the shared representation rather than the entire fused detection feature.

By restricting domain adaptation to the shared feature space, the detector is encouraged to learn domain-invariant object structures while avoiding over-alignment of heterogeneous SAR scattering patterns. Meanwhile, the scattering-specific expert paths are not directly constrained by the alignment loss, allowing them to preserve and compensate sensor-dependent responses caused by frequency band, resolution, clutter, shadow, and scattering-center variations. This shared-specific separation improves the robustness of few-shot cross-sensor SAR object detection.

\subsection{Optimization Objective}
\label{subsec:objective}

The overall training objective consists of the detection loss, the shared-representation alignment loss, the routing-domain auxiliary loss, and the expert balancing loss:
\begin{equation}
\mathcal{L}=\mathcal{L}_{det}+\lambda_{align}\mathcal{L}_{align}^{sh}+\lambda_{rd}\mathcal{L}_{rd}+\lambda_{bal}\mathcal{L}_{bal},
\label{eq:total_loss}
\end{equation}
where $\lambda_{align}$, $\lambda_{rd}$, and $\lambda_{bal}$ are trade-off coefficients.

The detection loss follows the YOLOv5~\cite{glenn_jocher_2022_734952} object detection objective:
\begin{equation}
\mathcal{L}_{det}=\mathcal{L}_{box}+\mathcal{L}_{obj}+\mathcal{L}_{cls},
\end{equation}
where $\mathcal{L}_{box}$, $\mathcal{L}_{obj}$, and $\mathcal{L}_{cls}$ denote the bounding-box regression loss, objectness loss, and classification loss, respectively. The detection loss is computed on both labeled source-domain samples and the few fully annotated target-domain samples.

The shared alignment loss $\mathcal{L}_{align}^{sh}$ is imposed only on the domain-shared representation. It encourages the detector to learn transferable object structures across sensors while avoiding over-alignment of scattering-specific expert features. The routing-domain loss $\mathcal{L}_{rd}$ is applied to the routing weights of the scattering-specific experts, encouraging the router to capture sensor-dependent scattering preferences.

To prevent routing collapse, we further introduce an expert balancing loss~\cite{lin2026yolo}. Let $\bar{\mathbf{r}}_l$ denote the average routing probability over a mini-batch at the $l$-th feature level:
\begin{equation}
\bar{\mathbf{r}}_l=\frac{1}{|\mathcal{B}|}\sum_{i\in\mathcal{B}}\mathbf{r}_l(x_i).
\end{equation}
The balancing loss is defined as
\begin{equation}
\mathcal{L}_{bal}=\sum_{l\in\mathcal{S}}\left\|\bar{\mathbf{r}}_l-\frac{1}{E}\mathbf{1}\right\|_2^2,
\label{eq:balance_loss}
\end{equation}
where $\mathcal{S}$ denotes the set of feature levels equipped with scattering-specific experts, $E$ is the number of experts, and $\mathbf{1}$ is an all-one vector. This loss encourages different experts to be sufficiently activated within a mini-batch, thus avoiding the degeneration where only a few experts dominate the routing process.

During inference, only the detector with the proposed shared-specific feature decomposition and soft-gated expert compensation is retained. The routing-domain classifier and all auxiliary losses are removed, and no target-domain labels are required.

\section{Experiments}
\label{sec:experiments}
\subsection{Datasets and Experimental Settings}
\label{subsec:datasets_settings}
\begin{figure}[t]
    \centering
    \includegraphics[width=\linewidth]{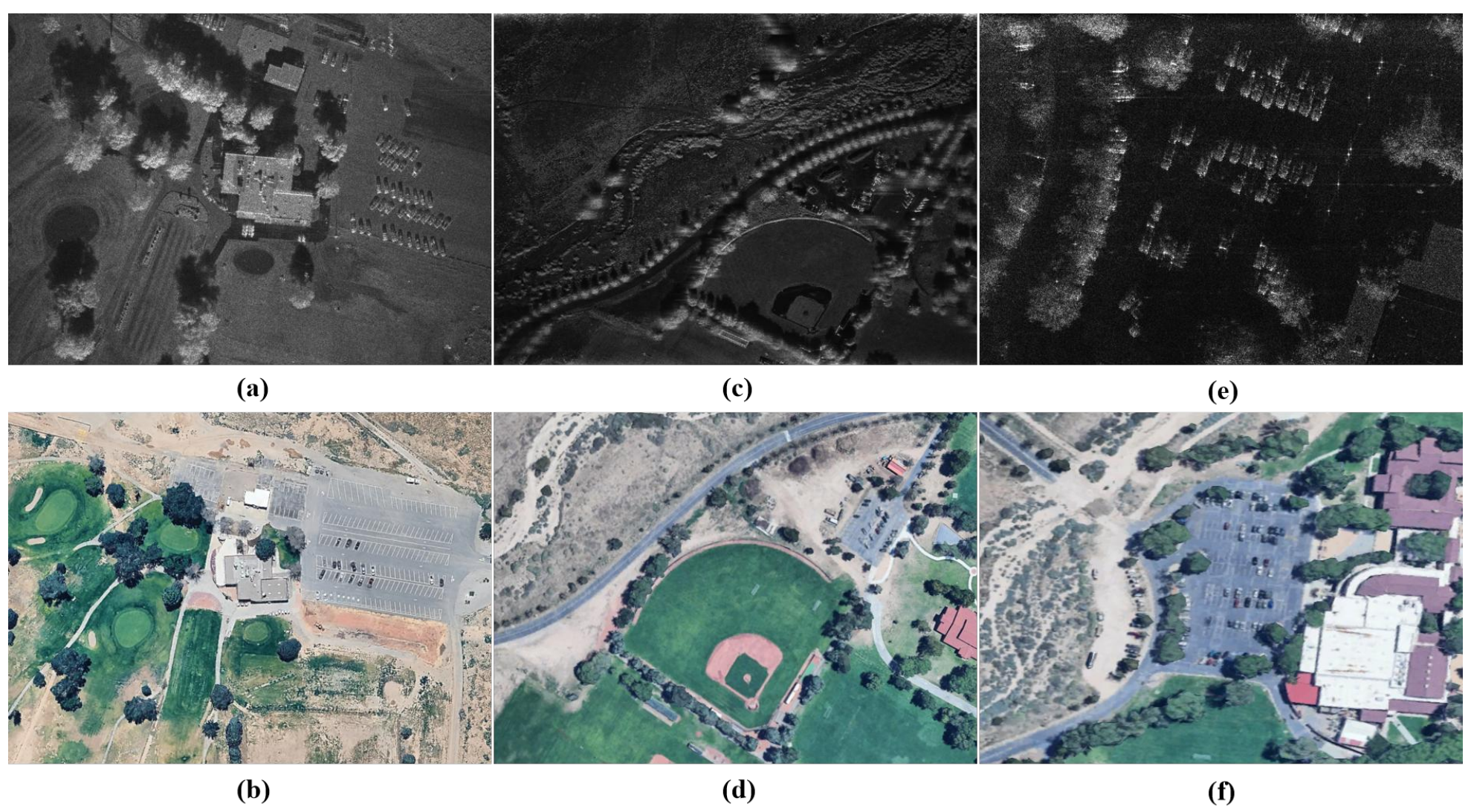}
    \caption{
    Examples of SAR images and corresponding optical scenes from the datasets used in our experiments. (a), (c), and (e) are SAR images from MiniSAR, FARAD-X, and FARAD-Ka, respectively. (b), (d), and (f) show the corresponding optical scenes on Google Earth.
    }
    \label{fig:dataset_examples}
\end{figure}

Experiments are conducted on the FARAD-X, FARAD-Ka~\cite{sandia_farad}, and MiniSAR~\cite{sandia_MiniSAR} vehicle datasets released by Sandia National Laboratories. As summarized in Table~\ref{tab:dataset_domain}, these datasets were acquired by different radar systems and exhibit evident discrepancies in frequency band, incidence angle, and scene clutter. Representative SAR and optical images are shown in Fig.~\ref{fig:dataset_examples}.
\begin{table}[t]
\centering
\caption{Imaging parameters and domain characteristics of the datasets.}
\label{tab:dataset_domain}

\resizebox{\columnwidth}{!}{
\begin{tabular}{c|ccc}
\toprule[1.0pt]

\multicolumn{1}{c|}{\textbf{Domain}}
& \textbf{FARAD-X} & \textbf{FARAD-Ka} & \textbf{MiniSAR} \\

\midrule[0.5pt]
Band & X & Ka & Ku \\
Time & 2015.10 & 2015.08 & 2005.05 \\
Bandwidth & 3 GHz & 5 GHz & 3 GHz \\
Center Freq. & 9.6 GHz & 35.6 GHz & 16.8 GHz \\
Angle & $26^\circ$--$34^\circ$ & $26^\circ$--$34^\circ$ & $26^\circ$--$29^\circ$ \\
Resolution & 0.1 m & 0.1 m & 0.1 m \\
Max Range & 12 km & 6 km & 8 km \\
Scene Type & Dense urban & Dense urban & Suburban \\
Location & \makecell[c]{Univ. of\\New Mexico} & \makecell[c]{Univ. of\\New Mexico} & \makecell[c]{Kirtland Air\\Force Base} \\
\bottomrule[1.0pt]
\end{tabular}
}
\end{table}
In this paper, we construct four bidirectional cross-sensor adaptation tasks: FARAD-X to MiniSAR, FARAD-Ka to MiniSAR, MiniSAR to FARAD-X, and MiniSAR to FARAD-Ka. The vehicle category is used as the only target class. Following the preprocessing strategy of SIVED~\cite{lin2023sived}, the original large-scale SAR images are cropped into $512\times512$ patches, and only patches containing at least one vehicle target are retained. 

For the forward tasks, the training split of FARAD-X or FARAD-Ka is used as the source domain, while 30 MiniSAR images form the target-domain candidate set and the remaining 70 images constitute a fixed test set. For the reverse tasks, the MiniSAR training split is used as the source domain, the FARAD training split forms the target-domain candidate set, and the corresponding validation and test splits are combined for evaluation. No target-domain evaluation image is used during adaptation. 

For the $K$-shot setting, $K$ fully annotated target images are selected from the candidate set for adaptation, where $K\in\{1,3,5,8,10\}$. The few-shot subsets are constructed in a nested manner~\cite{Wang2020FrustratinglySF}, i.e., the samples used in a lower-shot setting are included in the corresponding higher-shot setting under the same random seed. The unused images in the candidate set are not involved in training. 

All experiments are implemented based on the AsyFOD~\cite{gao2023asyfod} framework for fair comparison with representative few-shot domain adaptation baselines. The input SAR patches are resized to $512\times512$ during both training and testing. The detector is trained for 300 epochs using SGD with an initial learning rate of 0.01 and a batch size of 12, including 11 source-domain images and one target-domain image. All experts are used at each detection feature level. We set $\lambda_{rd}=0.1$, $\lambda_{bal}=0.002$, and $\tau=1$. The learnable coefficient $\alpha_l$ is initialized to 0.3, while $\beta_l$ is fixed to 1. All experiments are conducted on an NVIDIA GeForce RTX 5090 GPU, and performance is evaluated using mAP$_{50}$.

\subsection{Comparison with State-of-the-Art Methods}
\label{subsec:main_results}

\begin{table*}[t]
\centering
\caption{Comparison of different methods on bidirectional few-shot SAR domain adaptation tasks in terms of mAP$_{50}$ (\%).}
\label{tab:main_results}

\footnotesize
\setlength{\tabcolsep}{4.0pt}
\renewcommand{\arraystretch}{1.1}

\begin{tabular}{l|ccccc|ccccc}
\toprule[1.0pt]

\multicolumn{1}{c|}{
    \multirow[c]{2}{*}[-0.5ex]{\textbf{Methods}}
}
& \multicolumn{5}{c|}{\textbf{(a) FARAD-X $\rightarrow$ MiniSAR}}
& \multicolumn{5}{c}{\textbf{(b) FARAD-Ka $\rightarrow$ MiniSAR}} \\
\cmidrule(lr){2-6}
\cmidrule(lr){7-11}
&
\textbf{1-shot} & \textbf{3-shot} & \textbf{5-shot}
& \textbf{8-shot} & \textbf{10-shot}
&
\textbf{1-shot} & \textbf{3-shot} & \textbf{5-shot}
& \textbf{8-shot} & \textbf{10-shot} \\
\midrule[0.5pt]

Source-only~\cite{glenn_jocher_2022_734952}
& \multicolumn{5}{c|}{22.46}
& \multicolumn{5}{c}{25.77} \\

FSDet~\cite{Wang2020FrustratinglySF}
& 19.38 & 30.42 & 36.84 & 41.01 & 47.69
& 27.72 & 37.33 & 44.07 & 53.36 & 45.26 \\

SSDA-YOLO~\cite{Zhou2022SSDAYOLOSD}
& 23.13 & 28.48 & 39.04 & 40.08 & 39.45
& 29.17 & 39.68 & 46.67 & 46.56 & 50.69 \\

AcroFOD~\cite{Gao2022AcroFODAA}
& 27.35 & 36.20 & 51.81 & 61.29 & 64.53
& 30.98 & 41.30 & 55.56 & 64.08 & 66.18 \\

AsyFOD~\cite{gao2023asyfod}
& 29.09 & 36.67 & 49.60 & 56.58 & 63.37
& 37.43 & 36.90 & 52.78 & 60.91 & 58.62 \\

AsyFOD$^{\dagger}$~\cite{gao2023asyfod}
& 43.39 & 49.09 & 66.10 & 72.07 & 71.23
& 41.30 & 43.30 & 66.01 & 69.72 & 73.21 \\

\midrule[0.3pt]

\textbf{Our method}
& \textbf{47.19} & \textbf{51.19} & \textbf{68.98} & \textbf{76.06} & \textbf{73.58} 
& \textbf{47.26} & \textbf{51.60} & \textbf{69.32} & \textbf{75.72} & \textbf{75.42} \\

\midrule[1.0pt]

\multicolumn{1}{c|}{
    \multirow[c]{2}{*}[-0.5ex]{\textbf{Methods}}
}
& \multicolumn{5}{c|}{\textbf{(c) MiniSAR $\rightarrow$ FARAD-X}}
& \multicolumn{5}{c}{\textbf{(d) MiniSAR $\rightarrow$ FARAD-Ka}} \\
\cmidrule(lr){2-6}
\cmidrule(lr){7-11}
&
\textbf{1-shot} & \textbf{3-shot} & \textbf{5-shot}
& \textbf{8-shot} & \textbf{10-shot}
&
\textbf{1-shot} & \textbf{3-shot} & \textbf{5-shot}
& \textbf{8-shot} & \textbf{10-shot} \\
\midrule[0.5pt]

Source-only~\cite{glenn_jocher_2022_734952}
& \multicolumn{5}{c|}{19.42}
& \multicolumn{5}{c}{22.64} \\

FSDet~\cite{Wang2020FrustratinglySF}
& 8.22 & 42.94 & 53.97 & 56.97 & 58.05
& 15.62 & 42.28 & 48.48 & 54.56 & 49.77 \\

SSDA-YOLO~\cite{Zhou2022SSDAYOLOSD}
& 17.63 & 25.39 & 22.92 & 16.89 & 16.56
& 16.66 & 20.08 & 22.60 & 24.63 & 15.64 \\

AcroFOD~\cite{Gao2022AcroFODAA}
& 22.41 & 50.47 & 44.79 & 61.68 & 46.73
& 19.58 & 44.37 & 44.08 & 54.72 & 56.72 \\

AsyFOD~\cite{gao2023asyfod}
& 21.08 & 41.02 & 42.80 & 44.10 & 59.42
& 11.48 & 34.98 & 58.90 & 51.80 & 56.95 \\

AsyFOD$^{\dagger}$~\cite{gao2023asyfod}
& 23.94 & 52.85 & 55.44 & 62.75 & 64.45
& \textbf{26.26} & 52.24 & 61.98 & 64.62 & 64.57 \\

\midrule[0.3pt]

\textbf{Our method}
& \textbf{24.94} & \textbf{54.22} & \textbf{58.31} & \textbf{64.50} & \textbf{71.76}
& 23.94 & \textbf{55.80} & \textbf{62.00} & \textbf{72.51} & \textbf{69.77} \\
\bottomrule[1.0pt]
\end{tabular}

\vspace{2mm}

\begin{minipage}{\linewidth}
\footnotesize
\textit{Note:} The $K$-shot setting denotes the use of $K$ fully annotated images from the target domain. AsyFOD denotes the results reproduced using the released implementation, whereas AsyFOD$^{\dagger}$ denotes the reproduced results obtained after correcting the back-propagation of the asymmetric alignment loss.
\end{minipage}
\end{table*}

\begin{figure*}[!t]
    \centering
    \includegraphics[
        width=0.95\textwidth,
        height=0.82\textheight,
        keepaspectratio
    ]{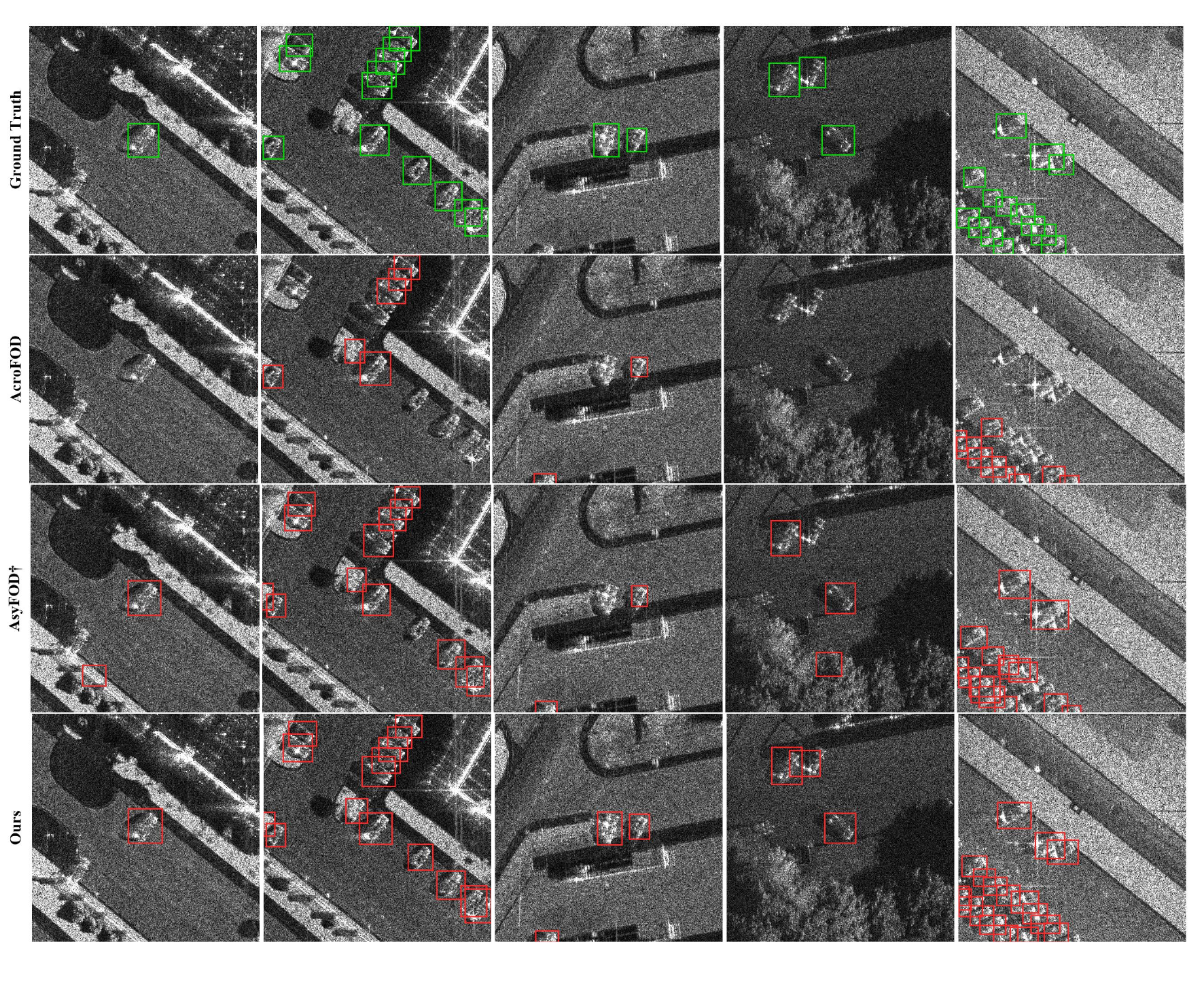}
    \caption{
        Qualitative comparison on MiniSAR test images under the 5-shot FARAD-X $\rightarrow$ MiniSAR adaptation task. Each column represents one test scene, while the rows from top to bottom correspond to the ground truth, AcroFOD, AsyFOD$^{\dagger}$, and the proposed method. Green and red boxes denote ground-truth annotations and detection results, respectively.
    }
    \label{fig:qualitative_results}
\end{figure*}

\begin{figure*}[t]
    \centering
    \includegraphics[width=0.98\textwidth]{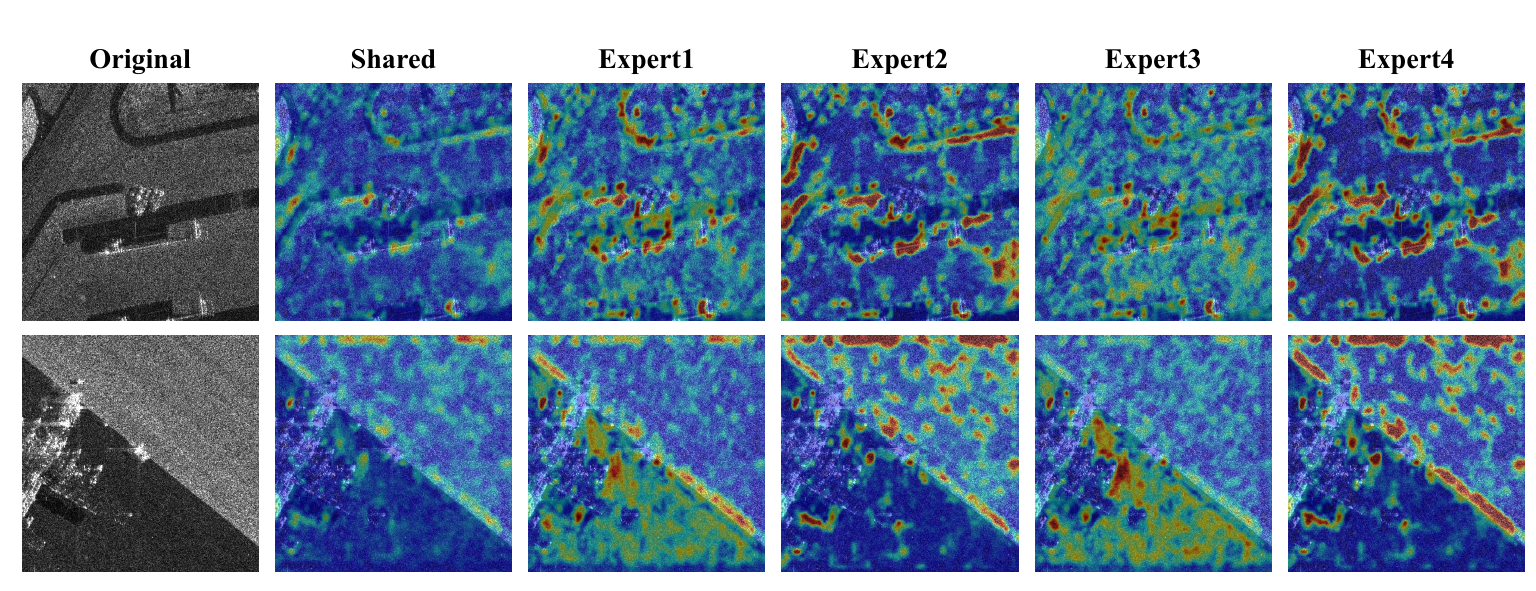}
    \caption{
    Visualization of P3 shared and expert activation responses on MiniSAR
    test images under the 5-shot FARAD-X $\rightarrow$ MiniSAR setting.
    From left to right: original image, shared path, and Experts 1--4.
    }
    \label{fig:expert_activation}
\end{figure*}
\begin{table*}[t]
\centering
\caption{Domain-wise router distribution statistics on the source and target image sets.
Each expert entry reports the source/target mean routing weights (\%).
TVD denotes the total variation distance between the source and target mean routing
distributions, and entropy denotes the normalized source/target routing entropy.}
\label{tab:router_domain}

\footnotesize
\setlength{\tabcolsep}{7.0pt}
\renewcommand{\arraystretch}{1.10}

\begin{tabular}{lccccccc}
\toprule
\textbf{Adaptation Task}
& \textbf{Level}& \textbf{Expert 1}& \textbf{Expert 2}& \textbf{Expert 3}& \textbf{Expert 4}& \textbf{TVD (\%)}& \textbf{Entropy} \\
\midrule\multirow{2}{*}{FARAD-X $\rightarrow$ MiniSAR}
& P3& 23.01/22.78& 28.47/29.02& 24.44/24.23& 24.08/23.97& 0.55& 0.9973/0.9966 \\
& P4& 24.30/24.94& 24.30/23.89& 26.62/28.08& 24.79/23.09& 2.10& 0.9993/0.9977 \\
\midrule\multirow{2}{*}{FARAD-Ka $\rightarrow$ MiniSAR}
& P3& 22.97/22.76& 26.73/26.87& 27.12/27.35& 23.18/23.02& 0.37& 0.9977/0.9973 \\
& P4& 25.89/25.50& 24.92/24.90& 25.30/25.16& 23.88/24.44& 0.56& 0.9996/0.9997 \\

\bottomrule
\end{tabular}
\end{table*}
To evaluate the effectiveness of the proposed method, we compare it with representative few-shot and domain adaptation baselines under the same SAR few-shot domain adaptation protocol. Source-only denotes the detector trained only with labeled source-domain images and directly evaluated on the target domain, which serves as the reference result without target-domain adaptation. Following AsyFOD, we implement the FSDet~\cite{Wang2020FrustratinglySF} baseline on YOLOv5 by freezing the first ten layers and fine-tuning the remaining parameters using the limited target-domain samples. SSDA-YOLO~\cite{Zhou2022SSDAYOLOSD} is included as a YOLO-based domain adaptation baseline and is adapted to the same few-shot target setting for comparison. In addition, AcroFOD~\cite{Gao2022AcroFODAA} and AsyFOD~\cite{gao2023asyfod} are selected as representative few-shot domain adaptive object detection methods. During reproduction, we found that the asymmetric alignment loss term in the released AsyFOD implementation was not involved in gradient back-propagation. We report both the results obtained using the released implementation, denoted as AsyFOD, and those obtained after correcting the gradient back-propagation, denoted as AsyFOD$^{\dagger}$. The basic detector, data augmentation strategy, optimizer, learning-rate schedule, and asymmetric alignment configuration are kept consistent with AsyFOD unless otherwise specified. For a fair comparison, all methods are evaluated using the same source-domain data, predefined target-domain few-shot subsets, and target-domain test sets.

Table~\ref{tab:main_results} reports the mAP$_{50}$ results on four cross-domain adaptation tasks under different few-shot settings. The upper panel presents the FARAD-X $\rightarrow$ MiniSAR and FARAD-Ka $\rightarrow$ MiniSAR results, while the lower panel presents the MiniSAR $\rightarrow$ FARAD-X and MiniSAR $\rightarrow$ FARAD-Ka results. Since Source-only does not use any target-domain samples for adaptation, its result is independent of the number of target shots. The results show that directly transferring the source-domain detector generally leads to limited target-domain performance because of the considerable discrepancy between heterogeneous SAR sensors. Most target-adapted methods, particularly AcroFOD and AsyFOD$^{\dagger}$, improve upon Source-only in the majority of settings, demonstrating the importance of using the limited target-domain annotations for cross-domain adaptation. 

Among the compared adaptation methods, AcroFOD and AsyFOD obtain strong performance by exploiting few-shot target annotations and source-domain knowledge. In particular, AsyFOD provides a competitive baseline by using target-guided source selection and asymmetric alignment. However, these methods mainly focus on sample selection or feature alignment, while sensor-dependent SAR scattering responses are not explicitly modeled. In contrast, the proposed method decomposes the detection representation into a mandatory shared path and soft-gated scattering-specific expert paths. The shared path provides transferable structural representations for source selection and domain alignment, while the scattering-specific experts preserve and adaptively compensate heterogeneous SAR responses.

As shown in the upper panel of Table~\ref{tab:main_results}, the proposed method achieves the best performance in all ten FARAD-to-MiniSAR settings. On FARAD-X $\rightarrow$ MiniSAR, our method improves upon AsyFOD$^{\dagger}$ by 3.80, 2.10, 2.88, 3.99, and 2.35 percentage points under the 1-, 3-, 5-, 8-, and 10-shot settings, respectively. On FARAD-Ka $\rightarrow$ MiniSAR, the corresponding improvements are 5.96, 8.30, 3.31, 6.00, and 2.21 percentage points. The lower panel further evaluates the proposed method in the reverse adaptation direction when MiniSAR is used as the source domain. Different from the forward tasks, only the 80 images in the MiniSAR training split are available as source-domain training data, whereas the forward tasks use 247 FARAD-X or 251
FARAD-Ka source training images. The reverse tasks therefore provide less source-domain supervision and more limited coverage of target and background variations, constituting a more challenging adaptation setting. On MiniSAR $\rightarrow$ FARAD-X, the proposed method outperforms AsyFOD$^{\dagger}$ under all five shot settings, with improvements of 1.00, 1.37, 2.87, 1.75, and 7.31 percentage points, respectively. On MiniSAR $\rightarrow$ FARAD-Ka, AsyFOD$^{\dagger}$ obtains the best 1-shot result of 26.26\%, whereas the proposed method achieves 23.94\%. Nevertheless, our method performs best in the remaining four settings and improves upon AsyFOD$^{\dagger}$ by 3.56, 0.02, 7.89, and 5.20 percentage points under the 3-, 5-, 8-, and 10-shot settings, respectively. The relatively lower result on the 1-shot MiniSAR $\rightarrow$ FARAD-Ka task suggests that, when both the source-domain training set and the target-domain adaptation set are extremely limited, the target distribution estimated from a single FARAD-Ka image may not be sufficiently representative. As more target-domain images become available, the routing preferences and shared target representation can be estimated more reliably, and the advantage of the proposed
scattering-specific compensation becomes more evident. These improvements demonstrate that aligning only the shared representations while preserving scattering-specific responses reduces the risk of over-aligning beneficial, sensor-dependent cues. These results validate the effectiveness of the proposed scattering-aware shared-specific feature decomposition for few-shot SAR domain adaptive object detection. Overall, the framework proves effective not only for adapting FARAD domains to MiniSAR, but also under the more challenging reverse setting with a markedly smaller MiniSAR source-domain training set.

Fig.~\ref{fig:qualitative_results} presents qualitative detection results on MiniSAR test images under the 5-shot FARAD-X $\rightarrow$ MiniSAR setting. Each column corresponds to the same test scene, while the rows from top to bottom show the ground truth, AcroFOD, AsyFOD$^{\dagger}$, and the proposed method, respectively. Compared with AcroFOD and AsyFOD$^{\dagger}$, the proposed method produces more reliable detection results in complex SAR scenes.

In the first column, AcroFOD fails to detect the isolated vehicle near the buildings, while AsyFOD$^{\dagger}$ produces additional false alarms in the surrounding background. In contrast, the proposed method correctly detects the target while suppressing background interference. In the third column, the vehicle adjacent to the building is particularly challenging because its scattering response can be confused with nearby strong background structures. Among the compared methods, only the proposed method successfully detects this target.

For the dense vehicle regions shown in the second and fifth columns, the proposed method also provides more complete detection results, indicating better robustness to densely distributed targets and cluttered SAR backgrounds. These observations suggest that the scattering-specific expert compensation helps preserve useful target-related scattering cues while reducing interference from clutter and strong background scatterers.

Fig.~\ref{fig:expert_activation} visualizes the P3 activation responses of the shared path and four experts on two representative MiniSAR test images. The shared path produces relatively smooth structural responses, whereas the experts exhibit more differentiated patterns. In these examples, Experts 1 and 3 show relatively stronger responses to target-adjacent shadow and surrounding context, while Experts 2 and 4 emphasize more localized scatterers, clutter edges, and structural boundaries. The experts therefore present an evident grouping tendency rather than four completely independent response modes. This observation indicates that the expert paths provide complementary scattering-related cues beyond the shared representation, and also offers qualitative support for the comparable performance of the two- and four-expert configurations reported in the subsequent parameter analysis.
To quantitatively complement the expert activation maps in Fig.~\ref{fig:expert_activation}, we further compare the router distributions of source- and target-domain images within the same trained model. As shown in Table~\ref{tab:router_domain}, all four experts maintain non-negligible routing weights, while their relative contributions vary across feature levels and domains. For FARAD-X $\rightarrow$ MiniSAR, a relatively larger domain-dependent variation is observed at P4, where the routing weight of Expert~3 increases from 26.62\% to 28.08\%, whereas that of Expert~4 decreases from 24.79\% to 23.09\%. The corresponding total variation distance reaches 2.10\%, which is larger than that at P3. In contrast, the FARAD-Ka
$\rightarrow$ MiniSAR model exhibits smaller routing variations, indicating a relatively stable expert combination with fine-grained domain-dependent adjustment. Meanwhile, the normalized routing
entropies remain close to one, demonstrating that the router performs soft expert reweighting rather than collapsing to a single expert. Together with the differentiated spatial responses in Fig.~\ref{fig:expert_activation}, these results show that the expert paths provide complementary representations whose relative contributions can be softly adjusted under cross-domain inputs without routing collapse.
\begin{figure}[t]
    \centering
    \includegraphics[width=\columnwidth]{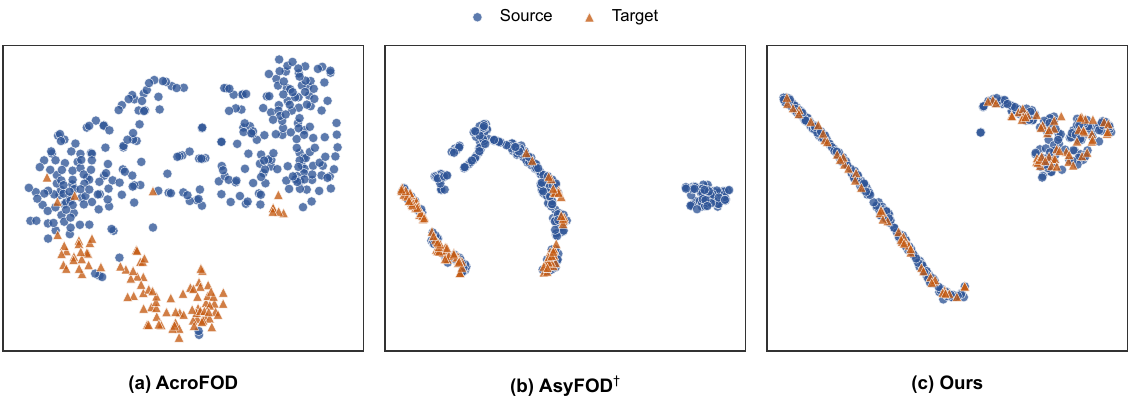}
    \caption{
    t-SNE visualization of source-domain and target-domain shared feature distributions under the 1-shot FARAD-Ka $\rightarrow$ MiniSAR adaptation task. 
    Blue circles and orange triangles denote source-domain and target-domain samples, respectively. 
    (a) AcroFOD. 
    (b) AsyFOD$^{\dagger}$. 
    (c) Ours. 
    The proposed method shows a closer source--target distribution in the shared feature space, indicating more effective transferable representation learning for few-shot SAR domain adaptation.
    }
    \label{fig:tsne_shared}
\end{figure}
We next analyze the adaptation behavior of the shared representation by visualizing the source-domain and target-domain feature distributions using t-SNE, as shown in Fig.~\ref{fig:tsne_shared}. For the proposed method, the visualized features are extracted from the shared path, which is used for target-guided source selection and domain alignment. For the baseline methods, we visualize the corresponding features used for alignment before the detection head. AcroFOD produces dispersed source and target features with limited overlap, indicating that insufficient alignment cannot effectively bridge the cross-sensor discrepancy. AsyFOD$^{\dagger}$ brings part of the source and target features closer, but the target features are highly concentrated under the 1-shot setting, while a large portion of source features still remain away from the target cluster. This suggests that direct alignment with extremely limited target samples may be biased toward a narrow target distribution. In contrast, the proposed method obtains a more compact and better-overlapped shared feature distribution, supporting the effectiveness of performing adaptation on the shared representation while leaving scattering-specific experts for residual compensation. As the number of target shots increases, the target-domain estimate becomes more reliable, and the alignment bias can be gradually alleviated. This observation supports our motivation that domain alignment should be mainly performed on shared representations, while scattering-specific expert paths are preserved to compensate sensor-dependent SAR responses.

\subsection{Ablation Study}
\label{subsec:ablation}
To evaluate the contribution of each component, we conduct ablation studies on FARAD-X $\rightarrow$ MiniSAR and FARAD-Ka $\rightarrow$ MiniSAR under different few-shot settings. AsyFOD$^{\dagger}$ is used as the baseline because it mainly performs adaptation through transferable feature alignment without explicitly modeling sensor-dependent scattering responses. The shared-only variant uses the proposed feature decomposition structure but retains only the shared path. We further evaluate the expert-only variant, fused-feature alignment, the shared-feature alignment loss $\mathcal{L}_{align}^{sh}$, the routing-domain loss $\mathcal{L}_{rd}$, and the expert balancing loss $\mathcal{L}_{bal}$.

As shown in Table~\ref{tab:ablation_components}, the shared-only variant obtains average mAP$_{50}$ values of 60.17\% and 59.81\% on the two tasks, which are close to the AsyFOD baseline. This shows that the shared path mainly preserves transferable features, while its improvement alone is limited. Applying alignment to the fused feature achieves average mAP$_{50}$ values of 58.05\% and 60.09\%, which are lower than those of the full model. This indicates that directly aligning the fused representation may suppress useful sensor-dependent responses retained by the expert paths. Moreover, removing $\mathcal{L}_{align}^{sh}$ reduces the average performance to 49.53\% and 52.44\%, confirming the importance of aligning the shared representation.

The expert-only variant achieves average mAP$_{50}$ values of 56.09\% and 61.50\%. Its performance is unstable on FARAD-X $\rightarrow$ MiniSAR, especially in the 5-shot setting. In comparison, the full model achieves 63.40\% and 63.86\% on the two tasks. These results show that shared features and expert features are complementary, and using either path alone cannot provide stable performance across different domains and shot numbers.

Without $\mathcal{L}_{rd}$, the average mAP$_{50}$ decreases to 61.39\% and 60.68\%, respectively. This suggests that domain-aware routing helps the experts learn sensor-related response patterns. Removing $\mathcal{L}_{bal}$ reduces the average performance to 62.05\% and 59.33\%. The large decrease in the 1-shot FARAD-Ka setting shows that the balancing loss helps prevent the router from relying too much on only a few experts.

Overall, the full model achieves the best average performance on both tasks. The results confirm that combining the shared path, scattering-specific experts, shared-feature alignment, routing-domain supervision, and expert balancing provides more effective and stable few-shot SAR domain adaptation.

\begin{table*}[t]
\centering
\caption{Ablation results of the proposed components in terms of
mAP$_{50}$ (\%).}
\label{tab:ablation_components}

\footnotesize
\setlength{\tabcolsep}{3.0pt}
\renewcommand{\arraystretch}{1.10}

\begin{tabular}{l|cccccc|cccccc}
\toprule[1.0pt]

\multicolumn{1}{c|}{
    \multirow[c]{2}{*}[-0.5ex]{\textbf{Methods}}
}
& \multicolumn{6}{c|}{
    \textbf{FARAD-X $\rightarrow$ MiniSAR}
}
& \multicolumn{6}{c}{
    \textbf{FARAD-Ka $\rightarrow$ MiniSAR}
} \\

\cmidrule(lr){2-7}
\cmidrule(lr){8-13}

& \textbf{1-shot} & \textbf{3-shot} & \textbf{5-shot} & \textbf{8-shot} & \textbf{10-shot} & \textbf{Avg.}
& \textbf{1-shot} & \textbf{3-shot} & \textbf{5-shot} & \textbf{8-shot} & \textbf{10-shot} & \textbf{Avg.} \\

\midrule[0.5pt]

AsyFOD$^{\dagger}$ baseline
& 43.39 & 49.09 & 66.10 & 72.07 & 71.23 & 60.38
& 41.30 & 43.30 & 66.01 & 69.72 & 73.21 & 58.71 \\

Fused-feature alignment
& 40.41 & 46.17 & 66.73 & 70.10 & 66.82 & 58.05
& 42.64 & 47.30 & 66.58 & 71.98 & 71.94 & 60.09 \\

Shared only
& 46.16 & 50.43 & 58.57 & 74.19 & 71.51 & 60.17
& 41.67 & 46.83 & 61.69 & 74.15 & 74.73 & 59.81 \\

w/o $\mathcal{L}_{align}^{sh}$
& 28.37 & 40.61 & 55.59 & 63.79 & 59.31 & 49.53
& 36.50 & 42.85 & 54.19 & 63.74 & 64.92 &  52.44\\

w/o $\mathcal{L}_{rd}$
& 46.32 & 50.17 & 63.81 & 73.92 & 72.71 & 61.39
& 41.05 & \textbf{51.67} & 68.01 & 69.55 & 73.10 & 60.68 \\

w/o $\mathcal{L}_{bal}$
& 45.23 & 51.14 & 68.12 & 74.59 & 71.19 & 62.05
& 28.64 & 51.53 & \textbf{69.32} & 72.55 & 74.63 & 59.33 \\

Expert only
& 44.60 & 49.56 & 42.96 & 71.24 & 72.10 & 56.09
& 43.70 & 48.32 & 66.77 & 74.83 & 73.86 & 61.50 \\

\midrule[0.3pt]

\textbf{Our method (full)}
& \textbf{47.19} & \textbf{51.19} & \textbf{68.98} & \textbf{76.06} & \textbf{73.58} & \textbf{63.40}
& \textbf{47.26} & 51.60 & \textbf{69.32} & \textbf{75.72} & \textbf{75.42} & \textbf{63.86} \\
\bottomrule[1.0pt]
\end{tabular}
\end{table*}

\subsection{Parameter Analysis}
\label{subsec:parameter_analysis}

We further analyze the influence of the number of scattering-specific experts. Tables~\ref{tab:num_experts_x} and~\ref{tab:num_experts_ka} report the mAP$_{50}$ results with different expert numbers on FARAD-X $\rightarrow$ MiniSAR and FARAD-Ka $\rightarrow$ MiniSAR, respectively. When only one expert is used, the model degenerates into a single scattering-specific transformation, which limits its ability to represent diverse SAR scattering variations. Using multiple experts usually provides a more flexible representation than a single expert, but the benefit is not strictly monotonic under the few-shot setting. E=2 and E=4 achieve comparable results, while E=8 may introduce redundant routing uncertainty.

It can be observed that $E=2$ and $E=4$ achieve comparable overall performance. Specifically, $E=4$ obtains the best average mAP$_{50}$ on FARAD-X $\rightarrow$ MiniSAR, while $E=2$ achieves a slightly higher average result on FARAD-Ka $\rightarrow$ MiniSAR. However, the difference between $E=2$ and $E=4$ is small. The four-expert configuration achieves the best average performance on FARAD-X $\rightarrow$ MiniSAR while remaining comparable to the two-expert configuration on FARAD-Ka $\rightarrow$ MiniSAR. When the number of experts is further increased to $E=8$, the performance does not consistently improve, suggesting that too many experts may introduce redundant routing uncertainty under the few-shot target-domain setting. Therefore, we set the number of scattering-specific experts to $E=4$ by default in the main experiments.

\begin{table}[t]
\centering
\caption{Effect of the number of scattering-specific experts on FARAD-X $\rightarrow$ MiniSAR in terms of mAP$_{50}$ (\%).}
\label{tab:num_experts_x}
\footnotesize
\setlength{\tabcolsep}{3.8pt}
\renewcommand{\arraystretch}{1.10}

\begin{tabular}{c|ccccc|c}
\toprule[1.0pt]
\multirow[c]{2}{*}[-0.5ex]{\textbf{Experts $E$}} & \multicolumn{5}{c|}{\textbf{FARAD-X $\rightarrow$ MiniSAR}} & \multirow[c]{2}{*}[-0.5ex]{\textbf{Avg.}} \\
\cmidrule(lr){2-6}
& \textbf{1-shot} & \textbf{3-shot} & \textbf{5-shot} & \textbf{8-shot} & \textbf{10-shot} & \\
\midrule[0.5pt]
1 & 44.97 & 52.21 & 68.35 & 70.49 & 68.77 & 60.96 \\
2 & 44.54 & \textbf{53.48} & 68.41 & 72.14 & 75.46 & 62.81 \\
4 & \textbf{47.19} & 51.19 & \textbf{68.98} & \textbf{76.06} & 73.58 & \textbf{63.40} \\
8 & 43.12 & 50.04 & 67.68 & 73.35 & \textbf{77.48} & 62.33 \\
\bottomrule[1.0pt]
\end{tabular}
\end{table}

\begin{table}[t]
\centering
\caption{Effect of the number of scattering-specific experts on FARAD-Ka $\rightarrow$ MiniSAR in terms of mAP$_{50}$ (\%).}
\label{tab:num_experts_ka}
\footnotesize
\setlength{\tabcolsep}{3.8pt}
\renewcommand{\arraystretch}{1.10}

\begin{tabular}{c|ccccc|c}
\toprule[1.0pt]
\multirow[c]{2}{*}[-0.5ex]{\textbf{Experts $E$}} & \multicolumn{5}{c|}{\textbf{FARAD-Ka $\rightarrow$ MiniSAR}} & \multirow[c]{2}{*}[-0.5ex]{\textbf{Avg.}} \\
\cmidrule(lr){2-6}
& \textbf{1-shot} & \textbf{3-shot} & \textbf{5-shot} & \textbf{8-shot} & \textbf{10-shot} & \\
\midrule[0.5pt]
1 & 44.17 & \textbf{54.18} & 70.73 & 76.77 & 75.52 & 64.27 \\
2 & \textbf{49.61} & 46.68 & \textbf{71.06} & \textbf{78.48} & 76.75 & \textbf{64.52} \\
4 & 47.26 & 51.60 & 69.32 & 75.72 & 75.42 & 63.86 \\
8 & 43.08 & 49.85 & 67.52 & 77.06 & \textbf{78.07} & 63.12 \\
\bottomrule[1.0pt]
\end{tabular}
\end{table}

\section{Conclusion}
\label{sec:conclusion}

We present a scattering-aware shared-specific feature decomposition framework for few-shot SAR domain adaptive object detection. Different from conventional feature-alignment methods that mainly learn domain-invariant representations, the proposed method considers that cross-domain SAR shifts contain both transferable target structures and sensor-dependent scattering responses. To avoid over-aligning useful scattering-specific cues, the proposed framework introduces a mandatory shared path for source selection and domain alignment, and several soft-gated scattering-specific experts for adaptive residual compensation. Routing-domain auxiliary supervision and expert balancing regularization are further used to encourage domain-sensitive expert weighting and stable expert utilization.

Experiments on four bidirectional cross-sensor adaptation tasks demonstrate the effectiveness of the proposed method in both adaptation directions. Compared with AsyFOD$^{\dagger}$, the proposed method improves the average mAP$_{50}$ by 3.02, 5.16, 2.86, and 2.87 percentage points on the four tasks, respectively, and achieves the best performance in 19 of the 20 evaluated task-shot settings. The reverse-task results further show that the proposed method remains effective when MiniSAR is used as the source domain and the more cluttered FARAD domains are used as target domains, although adaptation under the extremely limited 1-shot setting remains sensitive to the representativeness of the target samples. Ablation studies confirm that aligning shared representations while preserving scattering-specific expert compensation provides more stable few-shot cross-sensor adaptation.

\bibliographystyle{IEEEtran}
\bibliography{references}

\vfill

\end{document}